\documentclass[conference]{IEEEtran}
\IEEEoverridecommandlockouts
\usepackage{booktabs}
\usepackage{cite}
\usepackage{amsmath,amssymb,amsfonts}
\usepackage{algorithmic}
\usepackage{graphicx}
\usepackage{textcomp}
\usepackage{xcolor}
\usepackage{tikz}
\usepackage{subcaption}
\usetikzlibrary{arrows.meta, positioning, calc, fit, backgrounds, shapes.geometric, shapes.multipart}
\usepackage{xcolor}

\newcommand{\acc}[1]{#1\%}

\def\BibTeX{{\rm B\kern-.05em{\sc i\kern-.025em b}\kern-.08em
    T\kern-.1667em\lower.7ex\hbox{E}\kern-.125emX}}
\begin{document}

\title{Retrieval-Augmented Online Textual Memory for Episodic Memory Video Question Answering\\

\thanks{This research is supported by the project  PIACERI - PIAno di inCEntivi per la Ricerca di Ateneo 2024/2026 — Linea di Intervento I “Progetti di ricerca collaborativa”. We acknowledge ISCRA for awarding this project access to the LEONARDO supercomputer, owned by the EuroHPC Joint Undertaking, hosted by CINECA (Italy).
}
}

\author{
\IEEEauthorblockN{
Raffaele Calì$^1$,
Giuseppe Lando$^1$,
Rosario Forte$^1$,
Antonino Furnari$^1$
}
\IEEEauthorblockA{
$^1$ Department of Mathematics and Computer Science, University of Catania, Italy\\
\{raffaele.cali, giuseppe.lando\}@studium.unict.it,
rosario.forte@phd.unict.it,
antonino.furnari@unict.it
}
}

\maketitle


\begin{abstract}
Long egocentric videos pose a central challenge for multimodal language
models: the visual stream must be compressed into a memory representation that
can be searched efficiently while still preserving the evidence needed for
question answering. Building on an online textual-memory framework in which
first-person videos are converted into temporally grounded captions, this work
systematically studies the role of retrieval in episodic-memory video question
answering. We evaluate how different retrieval units, query formulations,
context budgets, and reranking strategies affect multiple-choice answering on
QAEgo4D-Closed, isolating retrieval as the main variable while keeping the
underlying generated memory fixed.
Our best configuration, \textsc{Parent}, matches fine-grained caption subchunks,
then provides their full parent captions to the reasoner. It achieves 56.4\%
accuracy, matching strong full-memory and retrieval baselines while exposing
only a small fraction of the textual memory at question time. More importantly,
our analysis shows that better temporal coverage does not necessarily translate
into better answers: increasing the retrieval budget improves temporal Hit@K and
Recall@K, but can reduce accuracy by introducing visually plausible distractors
and uncertain caption evidence. These findings suggest that retrieval-augmented
egocentric memory should not be optimized only for evidence recall, but also for
the reliability, specificity, and answer-discriminative value of the retrieved
context. The study motivates more structured episodic memories that move beyond
raw caption retrieval toward evidence-aware representations for long-form
egocentric reasoning.
\end{abstract}


\section{Introduction}

Egocentric video question answering requires a model to reason over what a
camera wearer saw, did, handled, and encountered across long first-person video
streams. This setting is closely related to artificial episodic memory, where an
agent should recall past events and use them to answer later queries
~\cite{tulving1972episodic,grauman2022ego4dworld3000hours,cvpr24_groundvqa}.
It is also central to wearable assistants, lifelogging, and embodied perception,
where questions may refer to objects, locations, actions, and state changes
observed long before the question is asked
~\cite{shen2024encode,wang2024lifelongmemory,yang2025egolifeegocentriclifeassistant}.
However, egocentric video exposes a fundamental scalability problem. A single
recording may contain many visually similar moments, repeated objects, partial
views, camera motion, occlusions, and small state changes. Directly processing
the video at question time is expensive, while reading an entire textual memory
can still exceed practical context budgets and introduce plausible distractors.

\begin{figure}[t]
    \centering
    \includegraphics[width=\linewidth]{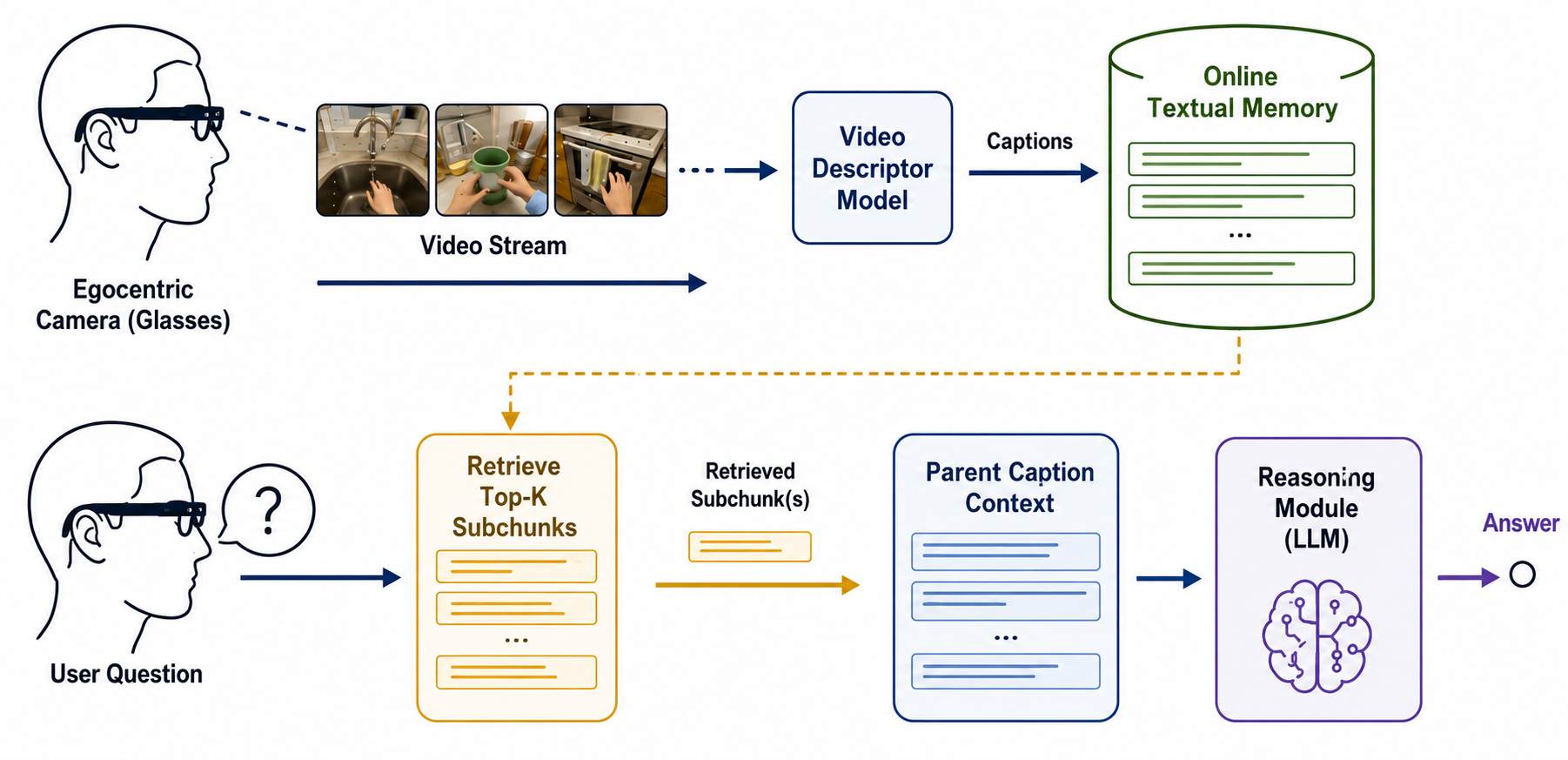}
    \caption{Overview of our retrieval-augmented textual-memory pipeline for egocentric video question answering. The egocentric video stream is processed online by a video descriptor model into captions, which are stored in an online textual memory. At question-answering time, the system uses the question and answer options to retrieve the most relevant subchunks, which are expanded to their parent captions and passed to the reasoner to predict the answer.}
    \label{fig:teaser}
\end{figure}

Recent work on Online Episodic Memory Video Question Answering (OEM-VQA)
addresses this issue by processing the visual stream sequentially and retaining a
compact memory that can be queried later, rather than reprocessing the full video
for every question
~\cite{di2025streamingvideoquestionansweringincontext,lando2025oemvqa}. Within
this direction, textual-memory approaches convert egocentric clips into language
descriptions, yielding memories that are compact, human-readable, and reusable
across multiple queries
~\cite{shen2024encode,wang2024lifelongmemory,yang2025egolifeegocentriclifeassistant,
lando2025oemvqa}. Prior work has shown that such memories can match dedicated
state-of-the-art systems on QAEgo4D-Closed while using substantially less memory
than approaches that retain visual features or cache large video representations
~\cite{cvpr24_groundvqa,di2025streamingvideoquestionansweringincontext,
lando2025oemvqa}. This makes textual memory an appealing substrate for scalable
and inspectable egocentric reasoning.

This paper builds on this textual-memory paradigm and studies a narrower but
crucial question: once the visual stream has been compressed into textual memory,
which parts of that memory should be provided to the reasoner at query time?
As illustrated in Fig.~\ref{fig:teaser}, our pipeline separates online memory
construction from question-time retrieval and reasoning. Video clips are first
converted into captions and stored in an online textual memory. When a question
arrives, the system retrieves relevant subchunks and expands them to their parent
captions, providing the reasoner with compact but contextual evidence.

Retrieval-augmented generation (RAG) was introduced to equip large language models with explicit non-parametric memory ~\cite{lewis2020retrieval}. In episodic video QA, however, retrieval does more
than reduce context length, it changes the evidence distribution seen by the
reasoner. A retrieval module must recover the relevant event, preserve enough
surrounding context to disambiguate the answer options, and avoid introducing
visually plausible but incorrect evidence from other moments in the same video.

We therefore isolate retrieval as the main object of study. All configurations
operate on the same generated textual memory and differ only in how memory
entries are indexed, retrieved, transformed, or passed to the reasoner. We compare
flat caption retrieval, subchunk retrieval, parent retrieval, hybrid
dense--lexical retrieval, and reranking strategies. Our strongest configuration
uses \textsc{Parent} retrieval: captions are split into fine-grained subchunks
for embedding and matching, but the corresponding full parent captions are
provided to the reasoner. This strategy reaches \acc{56.4} accuracy on
QAEgo4D-Closed, matching strong state-of-the-art baselines while using only
3{,}600 tokens per query on average.

The main finding of our study is that retrieval quality and answer accuracy do
not improve monotonically together. Increasing the number of retrieved captions
raises temporal Hit@K and Recall@K, but answer accuracy peaks earlier and then
declines. In long egocentric videos, additional retrieved context often contains
repeated objects, similar locations, uncertain descriptions, or actions that are
semantically close to the question but correspond to the wrong moment. As a
result, a reasoner may become less reliable even when the retrieved set covers
more of the annotated evidence window. This suggests that episodic-memory QA
should not optimize retrieval only for temporal coverage, but also for the
specificity and answer-discriminative value of the retrieved evidence.

Our contributions are threefold:
\begin{itemize}
    \item We provide a systematic retrieval-centred study for textual-memory
    OEM-VQA, explicitly building on prior training-free memory construction
    pipelines and isolating retrieval as the main variable.

    \item We compare multiple retrieval units, context budgets, and ranking
    strategies on QAEgo4D-Closed, showing that \textsc{Parent} retrieval provides
    the best trade-off between fine-grained matching and contextual reasoning.

    \item We analyze the mismatch between temporal retrieval metrics and final
    answer accuracy, showing that higher evidence coverage can introduce hard
    distractors and reduce the reliability of the final reasoner.
\end{itemize}
\section{Related Work}

\subsection{Egocentric Video Question Answering}

Episodic memory in egocentric vision concerns the ability to answer questions
about past first-person observations, such as where an object was placed, which
action was performed, or what happened before or after a given event
\cite{tulving1972episodic,grauman2022ego4dworld3000hours,cvpr24_groundvqa}. Ego4D
formalized this problem through the Natural Language Queries (NLQ) benchmark,
where models must localize the temporal segment containing the evidence needed
to answer a natural-language query \cite{grauman2022ego4dworld3000hours}.
Early work on episodic-memory QA further studied open-ended question answering
over egocentric videos \cite{9857465}, while Ground VQA introduced grounded
question answering in long egocentric videos and the QAEgo4D-Closed
multiple-choice setting \cite{cvpr24_groundvqa}. This formulation reduces the
ambiguity of free-form answer evaluation, but still requires models to identify
the relevant moment and distinguish the correct answer from plausible
distractors.

Recent work has increasingly focused on long-horizon and memory-based
egocentric understanding \cite{di2025streamingvideoquestionansweringincontext,ye2025mmega,wang2024lifelongmemory,yang2025egolifeegocentriclifeassistant,lando2025oemvqa}. MM-Ego studies egocentric multimodal LLMs for video QA
and highlights the difficulty of preserving visual details over long first-person
videos \cite{ye2025mmega}. Streaming video-language models such as
VideoLLM-online address online video understanding by processing video streams
incrementally rather than assuming that the full video can be reprocessed at
query time \cite{di2025streamingvideoquestionansweringincontext}. Complementary textual-memory approaches
decouple perception and reasoning by converting observed clips into language
descriptions and answering later questions over the accumulated memory
\cite{shen2024encode,wang2024lifelongmemory,yang2025egolifeegocentriclifeassistant,lando2025oemvqa}. Our work builds on this textual-memory paradigm, but
focuses specifically on retrieval: given an existing temporally grounded textual
memory, we study which evidence should be selected and provided to the reasoner
at query time.
\subsection{Video Retrieval-Augmented Generation}

Retrieval-augmented generation was introduced as a way to combine parametric
language models with explicit non-parametric memory \cite{lewis2020retrieval}.
In video understanding, this idea becomes more challenging because retrieval must
also handle temporal grounding and modality compression. DrVideo converts long
videos into document-like representations and retrieves relevant textual evidence
to support long-video understanding \cite{ma2025drvideo}. Video-RAG similarly
uses retrieval-augmented mechanisms to improve long-video comprehension by
selecting auxiliary visual-textual evidence before generation
\cite{ren2025videorag,luo2025videorag,alam2025openended}.
Our work is complementary to these video RAG systems. Rather than training or
designing a new video model, we isolate retrieval within an online textual-memory
pipeline whose captions have already been generated from the egocentric stream.
This allows us to study how retrieval units, query formulation, context budget,
and memory format affect downstream multiple-choice answering. In this setting,
retrieval is not only a way to reduce context length, it determines the evidence
distribution seen by the reasoner and can therefore improve or degrade answer
selection depending on the relevance and reliability of the retrieved captions.

\subsection{Memory Structures for Long Context Reasoning}

Textual memory is attractive for long-form egocentric reasoning because it is
compact, human-readable, easy to index, and reusable across multiple queries.
However, the structure of the memory determines what can be recovered at query
time. Long-context memory systems have shown that retrieval can benefit from
organizing information beyond independent chunks, for instance through
hippocampal-inspired memory and graph-based associations \cite{guo2024lightrag,gutierrez2024hipporag}.

This issue is especially important in egocentric video, where many questions are
not answered by object presence alone, but by temporal and relational context:
where an object appeared, when it was manipulated, what action preceded it, or
how it related to other objects and locations \cite{cvpr24_groundvqa,sun2026egograph,kim2026maegoqa}. For this reason, we analyze how retrieval over caption-based memories succeeds or fails when questions require temporally grounded object, action, and location evidence. Our results suggest that useful memory should preserve entities together with temporal anchors, actions, and locations~\cite{sun2026egograph,rajesh2026gsw}, but also keep the retrieved textual context compact. This indicates that caption-based episodic memories should optimize not only for evidence recall, but also for token efficiency, context specificity, and answer-discriminative value.
\section{Method}
\label{sec:method}

\begin{figure}[t]
    \centering
    \includegraphics[width=\linewidth]{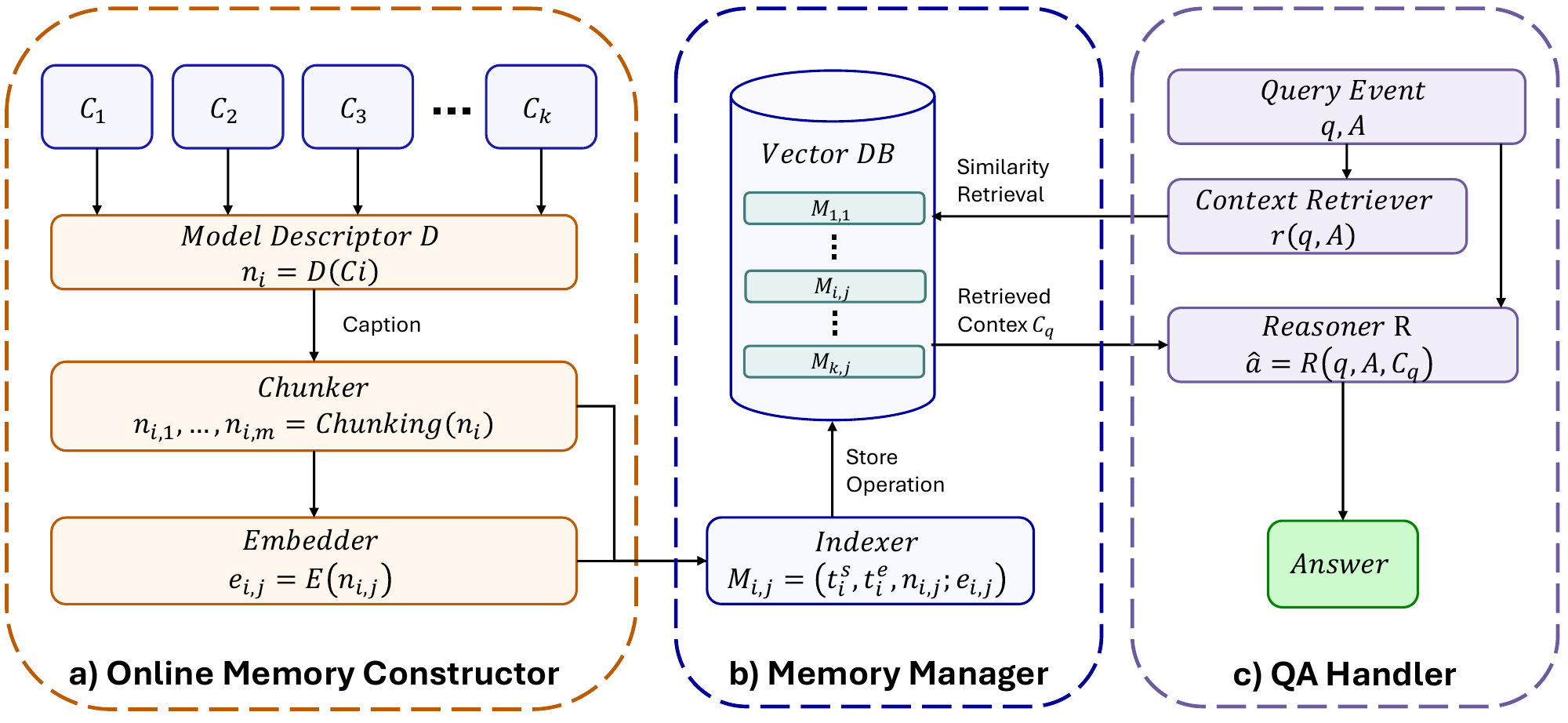}
    \caption{Overview of the proposed retrieval-augmented online textual-memory pipeline. The video stream is processed incrementally by a descriptor model into captions, which are then chunked, embedded, and stored in a vector database. At question-answering time, the query and candidate answers are used to retrieve relevant memory entries, which are provided to the reasoner to predict the final answer.}
    \label{fig:pipeline}
\end{figure}

Our retrieval-augmented formulation builds on the online textual-memory
pipeline introduced in~\cite{lando2025oemvqa}, where egocentric video clips are
processed incrementally into a compact textual memory and later queried by a
language-model reasoner. We extend this formulation by indexing the generated
captions and retrieving a query-specific subset of memory entries before
answering multiple-choice questions. The overall pipeline is shown in Fig.~\ref{fig:pipeline}. It is
organized into three main stages: an \emph{Online Memory Constructor}, a
\emph{Memory Manager}, and a \emph{QA Handler}.
Given an incoming video stream, clips are processed incrementally. For each new
clip, a multimodal descriptor model produces a natural-language caption. The
caption is then passed to a chunking module, which decomposes it into smaller
textual chunks, and each chunk is embedded by a text embedding model
(Fig.~\ref{fig:pipeline}(a)). 
Each chunk is stored with its temporal metadata and embedding in a vector database
(Fig.~\ref{fig:pipeline}(b)). At question-answering time, the input question and
the candidate answers are used to form a retrieval query. A context retriever
performs similarity search over the vector database and returns a retrieved
context. Finally, a language-model reasoner receives the question, the candidate
answers, and the retrieved context and predicts the final answer
(Fig.~\ref{fig:pipeline}(c)).
This formulation standardizes all memory operations around indexed textual
chunks. In contrast to treating a full caption as the only memory unit, our
pipeline explicitly separates caption generation, chunk construction,
embedding, indexing, retrieval, and reasoning.

\subsection{Online Memory Constructor}

Let $V$ denote an egocentric video stream. The stream is divided into a sequence
of temporally ordered clips
\begin{equation}
    V = \{c_1, c_2, \ldots, c_N\},
\end{equation}
where each clip $c_i$ corresponds to a temporal interval
$[t_i^{s}, t_i^{e}]$.

When a new clip $c_i$ becomes available, a multimodal descriptor model $D$
generates a natural-language caption:
\begin{equation}
    n_i = D(c_i).
\end{equation}
The caption $n_i$ describes the visual content observed in the temporal
interval $[t_i^{s}, t_i^{e}]$.
The caption is then passed to a chunking module, which decomposes it into a
set of textual chunks:
\begin{equation}
    \{n_{i,1}, n_{i,2}, \ldots, n_{i,m_i}\}
    =
    \operatorname{Chunking}(n_i),
\end{equation}
where $n_{i,j}$ denotes the $j$-th textual chunk generated from caption
$n_i$, and $m_i$ is the number of chunks produced for clip $c_i$. 

The output of the Online Memory Constructor is therefore a temporally grounded
set of textual chunks. Each chunk inherits the temporal interval of the clip
from which its parent caption was generated. This allows the memory to
preserve temporal information while operating at a finer textual granularity.

\subsection{Memory Manager}

The Memory Manager is responsible for embedding, indexing, and storing the
textual chunks produced by the Online Memory Constructor.

Each textual chunk $n_{i,j}$ is passed through a text embedding model $E$ 
\begin{equation}
    e_{i,j} = E(n_{i,j}),
\end{equation}
where $e_{i,j}$ is the dense vector representation of chunk $n_{i,j}$.

The indexer stores each chunk in the vector database as a memory entry:
\begin{equation}
    M_{i,j} =
    \left(t_i^{s}, t_i^{e}, n_{i,j}; e_{i,j}\right),
\end{equation}
where $t_i^{s}$ and $t_i^{e}$ are the start and end timestamps of the
corresponding clip, $n_{i,j}$ is the textual chunk, and $e_{i,j}$ is the
embedding used for indexing and similarity search.

After processing the first $i$ clips, the vector database contains the set of
indexed memory entries
\begin{equation}
    \mathcal{M}_i =
    \{M_{\ell,j} : 1 \leq \ell \leq i,\; 1 \leq j \leq m_\ell\}.
\end{equation}
The memory therefore grows incrementally as the video stream is processed. No
predefined upper bound on the number of memory entries is required.

For simplicity, when referring to the full memory after the stream has been
processed up to the point required by a question, we write
\begin{equation}
    \mathcal{M} =
    \{M_{i,j}\}_{i,j}.
\end{equation}

\subsection{QA Handler}

At question-answering time, the QA Handler receives a query event consisting of
a natural-language question $q$ and a set of candidate answers
$\mathcal{A} = \{a_1, a_2, \ldots, a_L\}$.
The context retriever builds a retrieval query from the question and from the candidate answers, defined as
\begin{equation}
    r(q,\mathcal{A}) = [q; a_1; \ldots; a_L].
\end{equation}
The notation $[q; a_1; \ldots; a_L]$ denotes the textual concatenation of the
question and all answer options.

The retrieval query is embedded using the same embedding model $E$ as
$e_q = E(r(q,\mathcal{A}))$, and the context retriever performs similarity search
over the vector database. Given a similarity function $\operatorname{sim}(\cdot,\cdot)$, the relevance score between the query and a
memory entry $M_{i,j}$ is defined as
\begin{equation}
    s(r(q,\mathcal{A}), M_{i,j})
    =
    \operatorname{sim}(e_q, e_{i,j}).
\end{equation}

The retrieved context is obtained by selecting the top-$K$ memory entries:
\begin{equation}
    \mathcal{C}_{q}
    =
    \operatorname{TopK}_{M_{i,j} \in \mathcal{M}}
    s(r(q,\mathcal{A}), M_{i,j}).
\end{equation}
Each retrieved entry contains both textual evidence and its associated temporal
interval, i.e. $\mathcal{C}_{q} = \{(t_i^{s}, t_i^{e}, n_{i,j})\}$ over the
retrieved entries.

Finally, the reasoner model $R$ receives the question, the candidate answers,
and the retrieved context, and predicts the final answer:
\begin{equation}
    \hat{a}
    =
    R(q,\mathcal{A},\mathcal{C}_{q}),
\end{equation}
where $\hat{a} \in \mathcal{A}$.

\begin{figure}
    \centering
    \includegraphics[width=1\linewidth]{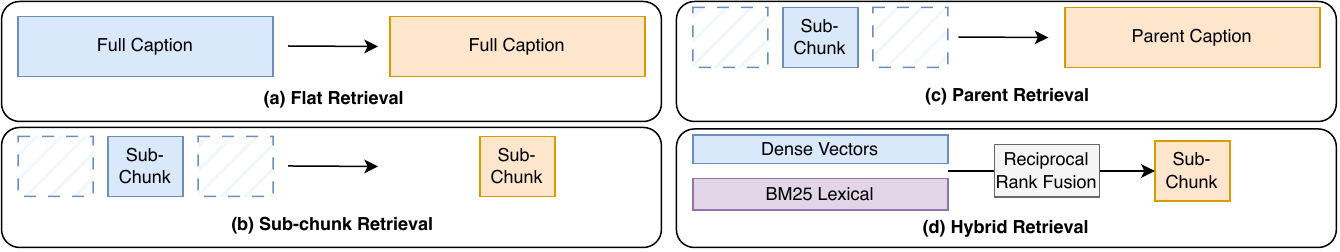}
    \caption{The proposed retrieval strategies.}
    \label{fig:retrieval}
\end{figure}

We consider different retrieval strategies (see Fig.~\ref{fig:retrieval}).

\subsection{Flat Caption Retrieval}

In the flat caption retrieval setting, each full caption $n_i$ is treated as a
single searchable unit instead of being decomposed into smaller chunks. In this
case, the memory entry associated with clip $c_i$ is
\begin{equation}
    M_i^{\mathrm{flat}}
    =
    \left(t_i^{s}, t_i^{e}, n_i; e_i\right),
\end{equation}
with $e_i = E(n_i)$. The retrieved context is obtained by selecting the top-$K$
full-caption entries by similarity:
\begin{equation}
    \mathcal{C}_{q}^{\mathrm{flat}}
    =
    \operatorname{TopK}_{M_i^{\mathrm{flat}}}
    \operatorname{sim}(E(r(q,\mathcal{A})), e_i).
\end{equation}

This strategy provides the reasoner with complete captions, preserving the
full local context generated for each clip. 

\subsection{Subchunk Retrieval}

Each caption $n_i$ is decomposed into smaller textual chunks
$n_{i,j}$, and each chunk is independently embedded and indexed as
$M_{i,j}$.

Retrieval is performed directly over the chunk-level vector database:
\begin{equation}
    \mathcal{C}_{q}^{\mathrm{sub}}
    =
    \operatorname{TopK}_{M_{i,j} \in \mathcal{M}}
    \operatorname{sim}(E(r(q,\mathcal{A})), e_{i,j}).
\end{equation}
The retrieved chunk-level context $\mathcal{C}_{q}^{\mathrm{sub}}$ is then passed
directly to the reasoner. This variant uses shorter and more focused evidence
snippets as the final retrieved context.

\subsection{Parent Retrieval}

The parent retrieval variant separates the unit used for retrieval from the
unit provided to the reasoner. Retrieval is performed over chunk-level entries
$M_{i,j}$, but each retrieved chunk is mapped back to its parent caption
$n_i$.

Let $\mathcal{U}_q$ denote the set of top-$K_s$ retrieved chunk-level entries:
\begin{equation}
    \mathcal{U}_q
    =
    \operatorname{TopK}_{M_{i,j} \in \mathcal{M}}
    \operatorname{sim}(E(r(q,\mathcal{A})), e_{i,j}).
\end{equation}

The parent context is obtained by collecting the full captions associated
with the retrieved chunks:
\begin{equation}
    \mathcal{C}_{q}^{\mathrm{parent}}
    =
    \operatorname{Unique}
    \{(t_i^{s}, t_i^{e}, n_i) : M_{i,j} \in \mathcal{U}_q\}.
\end{equation}
The reasoner then predicts the answer using the expanded parent context $\mathcal{C}_{q}^{\mathrm{parent}}$. This strategy uses fine-grained chunks for
retrieval while preserving the broader caption context for reasoning.

\subsection{Hybrid Retrieval}

We also evaluate a hybrid retrieval strategy that combines dense semantic
retrieval with lexical BM25 retrieval. Dense retrieval ranks memory entries
according to embedding similarity, while BM25 ranks them according to lexical
matching between the retrieval query and the textual chunks.

Given a memory entry $M_{i,j}$, let
$\rho_{\mathrm{dense}}(r(q,\mathcal{A}), M_{i,j})$ denote its rank according to
dense retrieval, and let
$\rho_{\mathrm{BM25}}(r(q,\mathcal{A}), M_{i,j})$ denote its rank according to
BM25. The two rankings are combined using reciprocal rank fusion:
\begin{equation}
\begin{aligned}
    s_{\mathrm{RRF}}(r(q,\mathcal{A}), M_{i,j})
    =
    &\frac{1}{k_0 +
    \rho_{\mathrm{dense}}(r(q,\mathcal{A}), M_{i,j})}
    \\
    &+
    \frac{1}{k_0 +
    \rho_{\mathrm{BM25}}(r(q,\mathcal{A}), M_{i,j})},
\end{aligned}
\end{equation}
where $k_0$ controls the contribution of lower-ranked entries.

The hybrid retrieved context is then defined as
\begin{equation}
    \mathcal{C}_{q}^{\mathrm{hybrid}}
    =
    \operatorname{TopK}_{M_{i,j} \in \mathcal{M}}
    s_{\mathrm{RRF}}(r(q,\mathcal{A}), M_{i,j}).
\end{equation}

As in parent retrieval, the fused top-$K$ subchunks are expanded to their parent captions before the reasoner produces the final prediction.

Hybrid retrieval is intended to combine the strengths of dense semantic search
and lexical matching. 





\subsection{Evaluation}

We report answer accuracy as the percentage of questions for which the predicted answer matches the ground-truth answer.
For retrieval diagnostics, we also report temporal Hit@K and Recall@K against the annotated evidence region of each question. Hit@$K$ measures whether at least one of the top-$K$ retrieved memory entries overlaps with the annotated evidence region for the question. Recall@$K$ measures how much of the annotated evidence region is covered by the temporal intervals of the top-$K$ retrieved entries.
By construction, Hit@K $\geq$ Recall@K, since Hit@K only requires a single overlapping entry while Recall@K measures the covered fraction of the evidence window.

Together, these diagnostics evaluate the temporal quality of the retrieval module, while answer accuracy measures the performance of the full retrieval-augmented reasoning pipeline.

\begin{table}[t]
\caption{Ablation of the proposed RAG configurations on QAEgo4D-Closed using a fixed retrieval budget of $K=12$ chunks.}
\label{tab:main}
\centering
\scriptsize
\renewcommand{\arraystretch}{1.08}
\begin{tabular}{l r r r}
\toprule
\textbf{System} & \textbf{Acc.} & \textbf{Hit@K} & \textbf{Recall@K} \\
\midrule
\textsc{Parent}        & \acc{56.4} & \acc{79.0} & \acc{70.7} \\
\textsc{Flat}          & \acc{55.6} & \acc{80.4} & \acc{72.3} \\
\textsc{Sub}           & \acc{54.2} & \acc{76.6} & \acc{67.5} \\
\textsc{Parent-Rerank} & \acc{53.8} & \acc{79.0} & \acc{70.7} \\
\textsc{Sub-Rerank}    & \acc{53.4} & \acc{82.4} & \acc{74.3} \\
\textsc{Hybrid}        & \acc{53.0} & \acc{78.8} & \acc{69.1} \\
\textsc{Parent-Time}   & \acc{52.8} & \acc{68.0} & \acc{58.8} \\
\bottomrule
\end{tabular}
\end{table}

\begin{figure}[t]
    \centering
    \includegraphics[width=\linewidth]{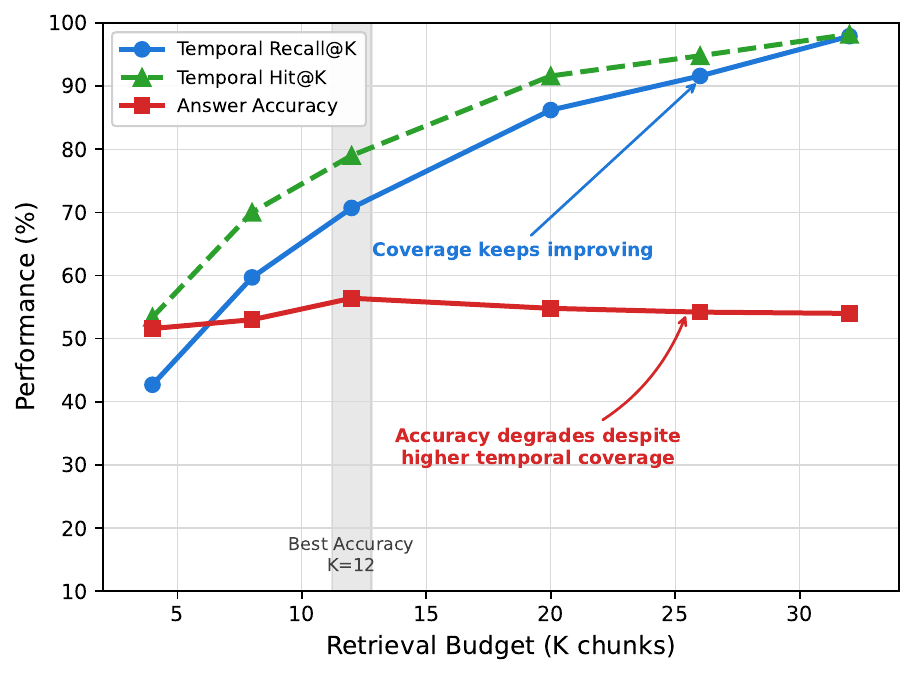}
    \caption{Retrieval budget curve for \textsc{Parent}. Increasing $K$ improves temporal coverage, but answer accuracy peaks at an intermediate retrieval budget.}
    \label{fig:t3}
\end{figure}

\section{Experimental Results and Analysis}
\label{sec:results}

This section reports the experimental results on the QAEgo4D-Closed~\cite{cvpr24_groundvqa} test set and analyzes how retrieval design affects both temporal evidence coverage and final answer accuracy.
All experiments use Qwen3-VL-8B~\cite{bai2025qwen3vltechnicalreport} as the descriptor model and Gemini 2.5 Flash~\cite{gemini} as the reasoner model, with temperature set to 0 and structured JSON decoding. 
Each video is processed in a streaming setting using non-overlapping clips of $15$ seconds. 

We use the captioning pipeline and resulting textual memory described in~\cite{lando2025oemvqa}, keeping the memory fixed across all experiments.
Dense retrieval uses \texttt{llama-3.2-nemoretriever-300m-embed-v1}~\cite{nvidia_llama32_nemoretriever_300m_embed_v1_2025}. Similarity is computed as cosine similarity, equivalently as dot product over normalized embeddings. The retrieval query is formed by concatenating the question with all answer options. Unless otherwise specified, we use a retrieval budget of $K=12$. 
For \textsc{Parent}-based methods, we retrieve $36$ subchunks and keep up to $K=12$ unique parent chunks as the final context before passing the context to the reasoner to ensure a fair comparison with other methods. \textsc{Hybrid} combines dense retrieval and BM25 using reciprocal rank fusion with $k_0=60$. \textsc{Parent-Time} applies a temporal cutoff to model the strictly online setting, retrieving only memory available up to the evidence time, serving as a diagnostic test. Reranking uses \texttt{nvidia/rerank-qa-mistral-4b}~\cite{nvidia_rerank_qa_mistral_4b} before parent expansion: \textsc{Parent-Rerank} reranks the same candidate pool as \textsc{Parent}, preserving the selected parents and temporal metric.

\begin{figure*}[t]
    \centering

    \begin{subfigure}[t]{0.48\textwidth}
        \centering
        \includegraphics[width=\linewidth]{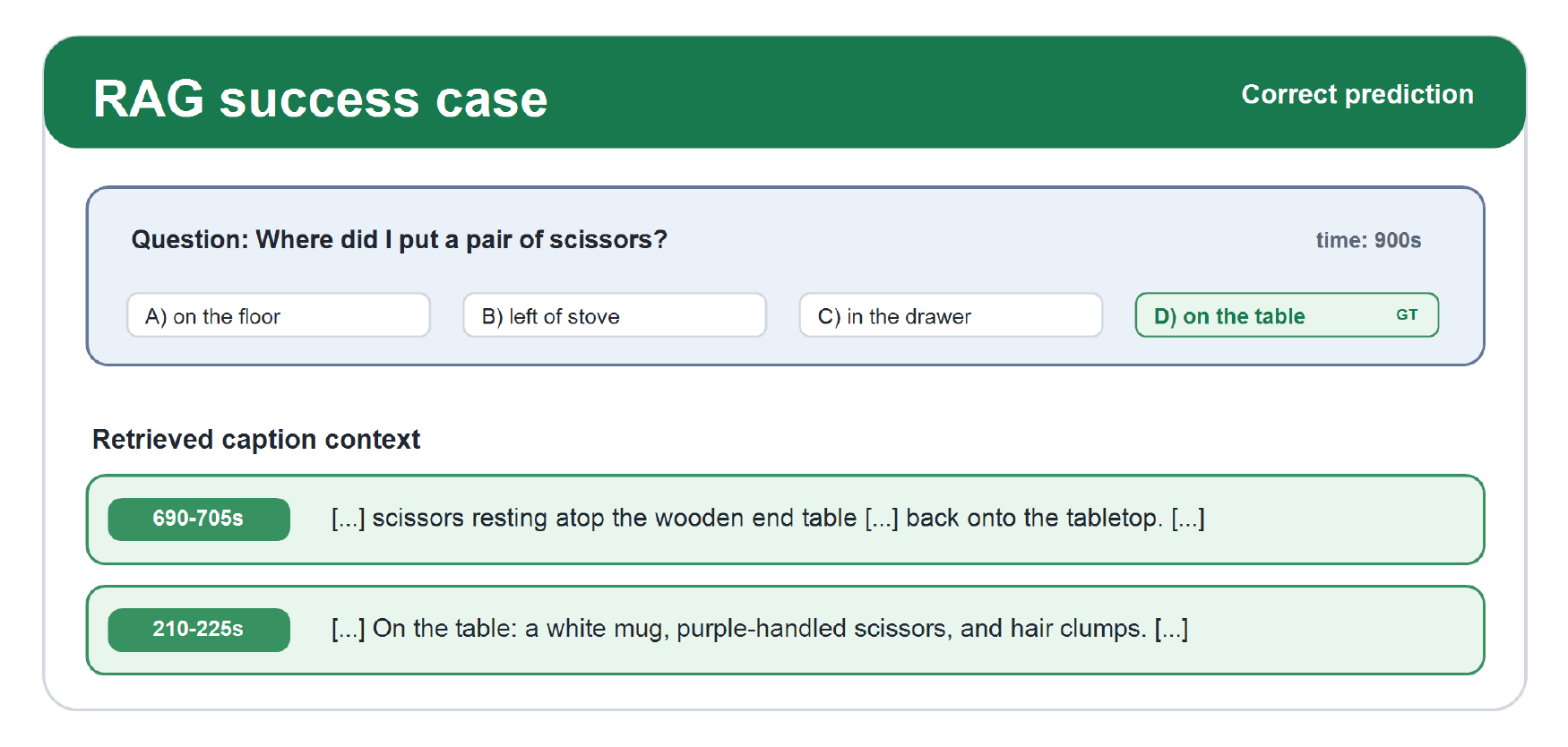}
        \label{fig:rag_success_case}
    \end{subfigure}
    \hfill
    \begin{subfigure}[t]{0.48\textwidth}
        \centering
        \includegraphics[width=\linewidth]{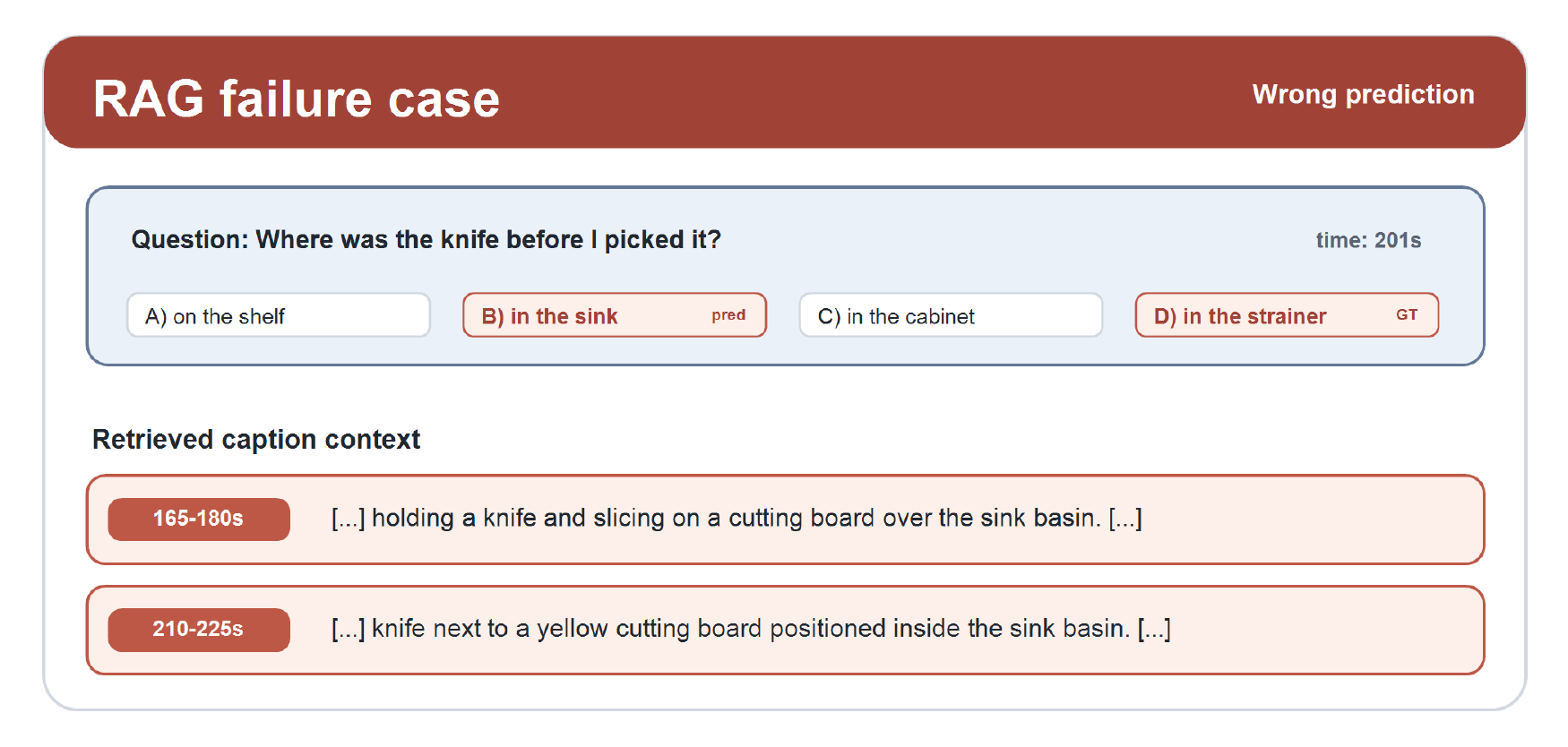}
        \label{fig:rag_failure_case}
    \end{subfigure}

    \caption{Qualitative examples of retrieval-augmented generation behavior.}
    \label{fig:rag_qualitative_cases}
\end{figure*}

\subsection{Analysis of Retrieval Strategies}

Table~\ref{tab:main} reports the ablation of the proposed RAG configurations using a fixed retrieval budget of $K=12$ retrieved units. 
The best result is obtained by \textsc{Parent}, which reaches \acc{56.4} accuracy, with a Hit@K of \acc{79.0} and a Recall@K of \acc{70.7}. 
The second-best configuration is \textsc{Flat}, which achieves a comparable accuracy of \acc{55.6} with slightly higher Hit@K and Recall@K.
This suggests that flat retrieval is already competitive when the retrieved captions contain sufficient context, while the parent strategy provides a better balance between precise matching and contextual grounding.

Interestingly, \textsc{Sub-Rerank} obtains the strongest retrieval scores among the tested configurations, with the highest Hit@K and Recall@K. 
However, this does not translate into the best final accuracy. 
This gap indicates that retrieving more relevant subchunks is not sufficient by itself, since the reasoner also needs enough surrounding information to correctly interpret the evidence and select the answer. 
In this sense, retrieval granularity plays a central role. 
Fine-grained subchunks are useful for locating the relevant moment, but using them alone can fragment the context available to the model. 
Conversely, parent captions preserve the temporal and semantic context around the matched evidence, leading to more robust answer generation. 

Overall, the results show that the most effective strategy is not simply the one with the highest retrieval metrics, but the one that best couples retrieval precision with readable context.

\subsection{Retrieval Budget and Temporal Coverage}

Figure~\ref{fig:t3} reports the retrieval budget curve for the best-performing \textsc{Parent} configuration. 
Increasing $K$ steadily improves temporal coverage: Hit@K rises from \acc{53.4} to \acc{98.2}, and Recall@K rises from \acc{42.7} to \acc{97.9}. 
Accuracy, however, follows a different trend. 
It peaks at \acc{56.4} for $K=12$, then decreases as additional captions are included.

This result highlights that temporal recall is useful but not sufficient for answer accuracy. 
In egocentric video, repeated rooms, objects, screens, tables, and actions create plausible hard negatives. 
A larger retrieved context can therefore cover more of the annotated evidence while also introducing distractors that make option selection harder for the reasoner.

\subsection{Comparison with State-of-the-Art Methods}

\begin{table}[t]
\caption{Comparison with state-of-the-art methods on QAEgo4D-Closed. 
For methods based on MLLMs, we report the number of tokens used at query time.}
\label{tab:sota_comparison}
\centering
\scriptsize
\renewcommand{\arraystretch}{1.08}
\begin{tabular}{l r r r}
\toprule
\textbf{Method} & \textbf{Acc.} & \textbf{Memory} & \textbf{Tokens} \\
\midrule
Ground VQA~\cite{cvpr24_groundvqa} & \acc{48.7} & 1.29 MB/min & -- \\
ReKV-LLaVAOneVision 0.5B~\cite{di2025streamingvideoquestionansweringincontext} & \acc{50.0} & 68.92 MB/min & 12{,}544 \\
ReKV-LLaVAOneVision 7B~\cite{di2025streamingvideoquestionansweringincontext} & \acc{56.0} & 329.64 MB/min & 12{,}544 \\
Lando et al. Gemini-based~\cite{lando2025oemvqa} & \acc{56.0} & \textbf{3.57 KB/min} & 10{,}880 \\
\midrule
\textsc{Parent} (Ours) & \textbf{\acc{56.4}} & 66.07 KB/min & \textbf{3{,}600} \\
\bottomrule
\end{tabular}
\end{table}

Table~\ref{tab:sota_comparison} compares our best-performing configuration with recent state-of-the-art methods on QAEgo4D-Closed~\cite{cvpr24_groundvqa}.
\textsc{Parent} reaches \acc{56.4} accuracy, which is on par with the best previously reported result, \acc{56.0}, obtained by the Gemini-based pipeline proposed in~\cite{lando2025oemvqa}.
The main advantage of our approach is its substantially lower query-time token usage. 
While the Gemini-based method of~\cite{lando2025oemvqa} uses the full textual memory, corresponding to 10{,}880 tokens at query time, our method requires only 3{,}600 tokens on average.
This corresponds to roughly 33\% of the full-memory input. 
Moreover, our memory footprint remains in the KB/min range, with 66.07 KB/min, whereas visual-memory and KV-cache-based methods such as Ground VQA~\cite{cvpr24_groundvqa} and ReKV~\cite{di2025streamingvideoquestionansweringincontext} operate in the MB/min range. 
Although these memory measurements are not directly equivalent across methods, since they represent different memory representations, they highlight that our textual-memory approach remains compact while being competitive with substantially heavier visual-memory baselines.

\subsection{Qualitative Examples}

Figure~\ref{fig:rag_qualitative_cases} shows representative success and failure cases of our RAG pipeline. 
In the success case, the retrieved captions explicitly mention the target object and its location, allowing the model to answer that the scissors were placed on the table correctly. 
In the failure case, the retrieved captions are semantically related to the query, but only provide coarse evidence about the knife being near the sink. 
As a result, the model predicts ``in the sink'' instead of the more precise ground-truth location, ``in the strainer.'' 
These examples suggest that our method is effective when retrieved captions contain explicit evidence, but suffers from missing or ambiguous fine-grained spatial details.
\section{Conclusion}

This study presented a training-free textual-memory pipeline for egocentric video question answering. 
Our best configuration, \textsc{Parent}, retrieves fine-grained caption subchunks and expands them to their parent captions, achieving \acc{56.4} accuracy on QAEgo4D-Closed while using a substantially smaller query-time context than full-memory reasoning.

Our analysis shows that temporal retrieval coverage does not necessarily translate into better answer accuracy. 
As the retrieval budget increases, Hit@K and Recall@K continue to improve, but accuracy peaks at an intermediate budget, since additional captions can introduce plausible distractors in long egocentric videos with repeated objects, locations, and actions.

These findings highlight the importance of retrieval design in textual-memory-based egocentric QA. 
Future work should move beyond treating captions as independent text units and develop evidence-aware episodic memories that preserve temporally grounded events, uncertainty, and answer-relevant support for downstream reasoning.

\bibliographystyle{IEEEtran}
\bibliography{Parts/references}

@article{tulving1972episodic,
  title={Episodic and semantic memory},
  author={Tulving, Endel and others},
  journal={Organization of memory},
  volume={1},
  number={381-403},
  pages={1},
  year={1972},
  publisher={New York}
}

@misc{grauman2022ego4dworld3000hours,
      title={Ego4D: Around the World in 3,000 Hours of Egocentric Video}, 
      author={{Ego4D Consortium}},
      year={2022},
      eprint={2110.07058},
      archivePrefix={arXiv},
      primaryClass={cs.CV},
      url={https://arxiv.org/abs/2110.07058}, 
}

@inproceedings{cvpr24_groundvqa,
  title={Grounded Question-Answering in Long Egocentric Videos},
  author={Di, Shangzhe and Xie, Weidi},
  booktitle={CVPR},
  year={2024}
}

@misc{di2025streamingvideoquestionansweringincontext,
  title        = {Streaming Video Question-Answering with In-context Video KV-Cache Retrieval},
  author       = {Shangzhe Di and Zhelun Yu and Guanghao Zhang and Haoyuan Li and Tao Zhong and Hao Cheng and Bolin Li and Wanggui He and Fangxun Shu and Hao Jiang},
  year         = {2025},
  howpublished = {arXiv preprint},
  eprint       = {2503.00540},
  archivePrefix= {arXiv},
  primaryClass = {cs.CV},
  url          = {https://arxiv.org/abs/2503.00540},
  note         = {arXiv:2503.00540}
}

@article{lando2025oemvqa,
  publtype={informal},
  author={Giuseppe Lando and Rosario Forte and Giovanni Maria Farinella and Antonino Furnari},
  title={How Far Can Off-the-Shelf Multimodal Large Language Models Go in Online Episodic Memory Question Answering?},
  year={2025},
  month={June},
  cdate={1748736000000},
  journal={CoRR},
  volume={abs/2506.16450},
  url={https://doi.org/10.48550/arXiv.2506.16450}
}

@inproceedings{shen2024encode,
  title={Encode-Store-Retrieve: Augmenting Human Memory through Language-Encoded Egocentric Perception},
  author={Shen, Junxiao and Dudley, John J and Kristensson, Per Ola},
  booktitle={2024 IEEE International Symposium on Mixed and Augmented Reality (ISMAR)},
  pages={923--931},
  year={2024},
  organization={IEEE}
}

@misc{wang2024lifelongmemory,
  title        = {LifelongMemory: Leveraging LLMs for Answering Queries in Long-form Egocentric Videos},
  author       = {Ying Wang and Yanlai Yang and Mengye Ren},
  year         = {2024},
  eprint       = {2312.05269},
  archivePrefix= {arXiv},
  howpublished = {arXiv preprint},
  primaryClass = {cs.CV},
  url          = {https://arxiv.org/abs/2312.05269},
  note={arXiv:2312.05269}
}

@misc{yang2025egolifeegocentriclifeassistant,
      title={EgoLife: Towards Egocentric Life Assistant}, 
      author={Jingkang Yang and Shuai Liu and Hongming Guo and Yuhao Dong and Xiamengwei Zhang and Sicheng Zhang and Pengyun Wang and Zitang Zhou and Binzhu Xie and Ziyue Wang and Bei Ouyang and Zhengyu Lin and Marco Cominelli and Zhongang Cai and Yuanhan Zhang and Peiyuan Zhang and Fangzhou Hong and Joerg Widmer and Francesco Gringoli and Lei Yang and Bo Li and Ziwei Liu},
      year={2025},
      eprint={2503.03803},
      archivePrefix={arXiv},
      primaryClass={cs.CV},
      url={https://arxiv.org/abs/2503.03803}, 
      note={arXiv:2503.03803}
}

@inproceedings{lewis2020retrieval,
  title={Retrieval-Augmented Generation for Knowledge-Intensive NLP Tasks},
  author={Lewis, Patrick and Perez, Ethan and Piktus, Aleksandra and Petroni, Fabio and Karpukhin, Vladimir and Goyal, Naman and K{\"u}ttler, Heinrich and Lewis, Mike and Yih, Wen-tau and Rockt{\"a}schel, Tim and Riedel, Sebastian and Kiela, Douwe},
  booktitle={Advances in Neural Information Processing Systems},
  volume={33},
  pages={9459--9474},
  year={2020}
}

@inproceedings{ye2025mmega,
  title={MM-Ego: Towards Building Egocentric Multimodal LLMs for Video QA},
  author={Ye, Hanrong and Zhang, Haotian and Daxberger, Erik and Chen, Lin and Lin, Zongyu and Li, Yanghao and Zhang, Bowen and You, Haoxuan and Xu, Dan and Gan, Zhe and Lu, Jiasen and Yang, Yinfei},
  booktitle={International Conference on Learning Representations},
  year={2025},
  url={https://arxiv.org/abs/2410.07177}
}

@misc{kim2026maegoqa,
  title={MA-EgoQA: Question Answering over Egocentric Videos from Multiple Embodied Agents},
  author={Kim, Kangsan and Yang, Yanlai and Kim, Suji and Yeo, Woongyeong and Lee, Youngwan and Ren, Mengye and Hwang, Sung Ju},
  year={2026},
  eprint={2603.09827},
  archivePrefix={arXiv},
  primaryClass={cs.CV},
  url={https://arxiv.org/abs/2603.09827},
  note={arXiv:2603.09827}
}

@misc{rajesh2026gsw,
  title={Beyond Fact Retrieval: Episodic Memory for RAG with Generative Semantic Workspaces},
  author={Rajesh, Shreyas and Holur, Pavan and Duan, Chenda and Chong, David and Roychowdhury, Vwani},
  year={2026},
  eprint={2511.07587},
  archivePrefix={arXiv},
  primaryClass={cs.AI},

  note={AAAI 2026 Oral; arXiv:2511.07587}
}

@misc{nvidia_llama32_nemoretriever_300m_embed_v1_2025,
  author       = {{NVIDIA}},
  title        = {{Llama 3.2 NeMo Retriever Embedding 300M v1}},
  howpublished = {NVIDIA NIM model card},
  year         = {2025},
  note         = {\url{https://docs.api.nvidia.com/nim}}
}

@misc{nvidia_rerank_qa_mistral_4b,
  author       = {{NVIDIA}},
  title        = {{NVIDIA Retrieval QA Text Reranking Mistral 4B}},
  howpublished = {NVIDIA NIM model card},
  note         = {\url{https://docs.api.nvidia.com/nim/reference/nvidia-rerank-qa-mistral-4b}}
}

@misc{sun2026egograph,
  title={EgoGraph: Temporal Knowledge Graph for Egocentric Video Understanding},
  author={Sun, Shitong and Han, Ke and Huang, Yukai and Cai, Weitong and Song, Jifei},
  year={2026},
  eprint={2602.23709},
  archivePrefix={arXiv},
  primaryClass={cs.CV},
  url={https://arxiv.org/abs/2602.23709},
  note={arXiv:2602.23709}
}

@misc{ren2025videorag,
  title={VideoRAG: Retrieval-Augmented Generation with Extreme Long-Context Videos},
  author={Ren, Xubin and Xu, Lingrui and Xia, Long and Wang, Shuaiqiang and Yin, Dawei and Huang, Chao},
  year={2025},
  eprint={2502.01549},
  archivePrefix={arXiv},
  primaryClass={cs.IR},
  url={https://arxiv.org/abs/2502.01549},
  note={arXiv:2502.01549}
}

@inproceedings{luo2025videorag,
  title={Video-RAG: Visually-aligned Retrieval-Augmented Long Video Comprehension},
  author={Luo, Yongdong and Zheng, Xiawu and Li, Guilin and Yin, Shukang and Lin, Haojia and Fu, Chaoyou and Huang, Jinfa and Ji, Jiayi and Chao, Fei and Luo, Jiebo and Ji, Rongrong},
  booktitle={Advances in Neural Information Processing Systems},
  year={2025},
  url={https://arxiv.org/abs/2411.13093}
}

@inproceedings{ma2025drvideo,
  title={DrVideo: Document Retrieval Based Long Video Understanding},
  author={Ma, Ziyu and Gou, Chenhui and Shi, Hengcan and Sun, Bin and Li, Shutao and Rezatofighi, Hamid and Cai, Jianfei},
  booktitle={Proceedings of the IEEE/CVF Conference on Computer Vision and Pattern Recognition},
  pages={18936--18946},
  year={2025}
}

@misc{alam2025openended,
  title={Open-Ended and Knowledge-Intensive Video Question Answering},
  author={Alam, Md Zarif Ul and Zamani, Hamed},
  year={2025},
  eprint={2502.11747},
  archivePrefix={arXiv},
  primaryClass={cs.IR},
  url={https://arxiv.org/abs/2502.11747},
  note={arXiv:2502.11747}
}

@misc{guo2024lightrag,
  title={LightRAG: Simple and Fast Retrieval-Augmented Generation},
  author={Guo, Zirui and Xia, Lianghao and Yu, Yanhua and Ao, Tu and Huang, Chao},
  year={2024},
  eprint={2410.05779},
  archivePrefix={arXiv},
  primaryClass={cs.IR},
  url={https://arxiv.org/abs/2410.05779},
  note={arXiv:2410.05779}
}

@inproceedings{gutierrez2024hipporag,
  title={HippoRAG: Neurobiologically Inspired Long-Term Memory for Large Language Models},
  author={Guti{\'e}rrez, Bernal Jim{\'e}nez and Shu, Yiheng and Gu, Yu and Yasunaga, Michihiro and Su, Yu},
  booktitle={Advances in Neural Information Processing Systems},
  year={2024},
  url={https://arxiv.org/abs/2405.14831}
}

@INPROCEEDINGS{9857465, author={Bärmann, Leonard and Waibel, Alex}, booktitle={2022 IEEE/CVF Conference on Computer Vision and Pattern Recognition Workshops (CVPRW)}, title={Where did I leave my keys? — Episodic-Memory-Based Question Answering on Egocentric Videos}, year={2022}, volume={}, number={}, pages={1559-1567}, doi={10.1109/CVPRW56347.2022.00162}}

@misc{gemini,
      title={Gemini 2.5: Pushing the Frontier with Advanced Reasoning, Multimodality, Long Context, and Next Generation Agentic Capabilities}, 
      author={Gemini Team},
      year={2025},
      eprint={2507.06261},
      archivePrefix={arXiv},
      primaryClass={cs.CL},
      url={https://arxiv.org/abs/2507.06261}, 
}

@misc{bai2025qwen3vltechnicalreport,
      title={Qwen3-VL Technical Report}, 
      author={Shuai Bai and Yuxuan Cai and Ruizhe Chen and Keqin Chen and Xionghui Chen and Zesen Cheng and Lianghao Deng and Wei Ding and Chang Gao and Chunjiang Ge and Wenbin Ge and Zhifang Guo and Qidong Huang and Jie Huang and Fei Huang and Binyuan Hui and Shutong Jiang and Zhaohai Li and Mingsheng Li and Mei Li and Kaixin Li and Zicheng Lin and Junyang Lin and Xuejing Liu and Jiawei Liu and Chenglong Liu and Yang Liu and Dayiheng Liu and Shixuan Liu and Dunjie Lu and Ruilin Luo and Chenxu Lv and Rui Men and Lingchen Meng and Xuancheng Ren and Xingzhang Ren and Sibo Song and Yuchong Sun and Jun Tang and Jianhong Tu and Jianqiang Wan and Peng Wang and Pengfei Wang and Qiuyue Wang and Yuxuan Wang and Tianbao Xie and Yiheng Xu and Haiyang Xu and Jin Xu and Zhibo Yang and Mingkun Yang and Jianxin Yang and An Yang and Bowen Yu and Fei Zhang and Hang Zhang and Xi Zhang and Bo Zheng and Humen Zhong and Jingren Zhou and Fan Zhou and Jing Zhou and Yuanzhi Zhu and Ke Zhu},
      year={2025},
      eprint={2511.21631},
      archivePrefix={arXiv},
      primaryClass={cs.CV},
      url={https://arxiv.org/abs/2511.21631}, 
}

\end{document}